\documentclass{article} % For LaTeX2e
\usepackage{iclr2019_conference,times}
\iclrfinalcopy
\usepackage[utf8]{inputenc} % allow utf-8 input
\usepackage[T1]{fontenc}    % use 8-bit T1 fonts
\usepackage{hyperref}       % hyperlinks
\usepackage{url}            % simple URL typesetting
\usepackage{booktabs}       % professional-quality tables
\usepackage{amsfonts}       % blackboard math symbols
\usepackage{nicefrac}       % compact symbols for 1/2, etc.
\usepackage{microtype}      % microtypography
\usepackage{color}
\usepackage{amsmath, amsthm, amssymb}
\usepackage{bm}
\usepackage{graphicx}
\usepackage{multirow}
\usepackage{tabularx}
\usepackage{booktabs}
\usepackage[ruled,vlined]{algorithm2e}
\usepackage{algorithmic}
\usepackage{float}
\usepackage{wrapfig}
\restylefloat{table}

\newcommand{\hide}[1]{}

\newcommand\todo[1]{\textcolor{red}{#1}}
\newcommand{\yuandong}[1]{\textcolor{blue}{[Yuandong: #1]}}

\newcommand{\eg}{\textit{e}.\textit{g}. }

\renewcommand{\cite}{\citep}

\def\eg{e.g. }

\def\etal{\emph{et al}. }

\title{Learning and Planning with a Semantic Model}

\author{Yi Wu \\
	UC Berkeley\\
	\texttt{jxwuyi@gmail.com} \\
	\And
	Yuxin Wu \\
	Facebook AI Research\\
	\texttt{yuxinwu@fb.com}\\
	\And
	Aviv Tamar \\
	UC Berkeley \\
	\texttt{avivt@berkeley.edu} \\
	\And
	Stuart Russell \\
	UC Berkeley\\
	\texttt{russell@berkeley.edu}\\
	\And
	Georgia Gkioxari \\
	Facebook AI Research \\
	\texttt{gkioxari@fb.com}\\
	\And
	Yuandong Tian \\
	Facebook AI Research \\
	\texttt{yuandong@fb.com}
}

\begin{document}

\maketitle
% this paper describes progresses on this challenge in the context of man-made environments. 
\begin{abstract}
Building deep reinforcement learning agents that can generalize and adapt to unseen environments remains a fundamental challenge for AI. This paper describes progresses on this challenge in the context of man-made environments, which are visually diverse but contain intrinsic semantic regularities. We propose a hybrid model-based and model-free approach, \emph{LEArning and Planning with Semantics (LEAPS)}, consisting of a multi-target sub-policy that acts on visual inputs, and a Bayesian model over semantic structures.
When placed in an unseen environment, the agent plans with the semantic model to make high-level decisions, proposes the next sub-target for the sub-policy to execute, and updates the semantic model based on new observations. We perform experiments in visual navigation tasks using House3D, a 3D environment that contains diverse human-designed indoor scenes with real-world objects. LEAPS outperforms strong baselines that do not explicitly plan using the semantic content.
\end{abstract}

\section{Introduction}
\label{sec:intro}

Deep reinforcement learning (DRL) has undoubtedly witnessed strong achievements in recent years~\cite{silver2016mastering,mnih2015human,levine2016}.
However, training an agent to solve tasks in a new unseen scenario, usually referred to as its \emph{generalization ability}, remains a challenging problem~\cite{geffner2018model,lake2017building}.
In model-free RL, the agent is trained to reactively make decisions from the observations, e.g., first-person view, via a black-box policy approximator. However the generalization ability of agents trained by model-free RL is limited, and is even more evident on tasks that require extensive planning~\cite{tamar2016value,kansky2017schema}.  On the other hand, model-based RL learns a dynamics model, predicting the next observation when taking an action. With the model, sequential decisions can be made via planning. However, learning a model for complex tasks and with high dimensional observations, such as images, is challenging. Current approaches for learning action-conditional models from video are only accurate for very short horizons~\cite{finn2017deep,ebert2017self,oh2015action}. Moreover, it is not clear how to efficiently adapt such models to changes in the domain.

In this work, we aim to improve the generalization of RL agents in domains that involve high-dimensional observations. Our insight is that in many realistic settings, building a pixel-accurate model of the dynamics is not necessary for planning high-level decisions. There are semantic structures and properties that are shared in real-world man-made environments. For example, rooms in indoor scenes are often arranged by their mutual functionality (\eg, bathroom next to bedroom, dining room next to kitchen). Similarly, objects in rooms are placed at locations of practical significance (\eg, nightstand next to bed, chair next to table). Humans often make use of such structural priors when exploring a new scene, or when making a high-level plan of actions in the domain. However, pixel-level details are still necessary for carrying out the high-level plan. For example, we need high-fidelity observations to locate and interact with objects, open doors, etc.

Based on this observation, we propose a hybrid framework, \emph{LEArning and Planning with Semantics (LEAPS)}, which consists of a model-based component that works on the semantic level to pursue a high-level target, and a model-free component that executes the target by acting on pixel-level inputs. Concretely, 
we (1) train model-free multi-target subpolicies in the form of neural networks that take the first-person views as input and sequentially execute sub-targets towards the final goal; (2) build a \emph{semantic} model in the form of a latent variable model that only takes \emph{semantic signals}, i.e., low-dimensional binary vectors, as input and is dynamically updated to plan the next sub-target. 
LEAPS has following advantages: (1) via model-based planning, generalization ability is improved; (2) by learning the \emph{prior} distribution of the latent variable model, we capture the semantic consistency among the environments; (3) the semantic model can be efficiently updated by posterior inference when the agent is exploring the unseen environment, which is effective even with very few exploration experiences thanks to the Bayes rule; and (4) the semantic model is lightweight and fully interpretable.

Our approach requires observations that are composed of both pixel-level data and a list of semantic properties of the scene. In general, automatically extracting high-level semantic structure from data is  difficult. As a first step, in this work we focus on domains where obtaining semantics is easy. In particular, we consider real-world environments for which strong object detectors are available~\cite{he2017mask}. An example of such environments is House3D which contains 45k real-world 3D scenes~\cite{wu2018building}. House3D provides a diverse set of scene layouts, object types, sizes and connectivity, which all conform to a consistent ``natural'' semantics. Within these complex scenes, we tackle navigation tasks within novel indoor scenes. Note that this problem is extremely challenging as the agent needs to reach far-away targets which can only be completed effectively if it can successfully reason about the overall structure of the new scenario. 
Lastly, we emphasize that although we consider navigation as a concrete example in this work, our approach is general and can be applied to other tasks for which semantic structures and signals are available

\hide{
Our extensive experiments show that our LEAPS framework can significantly improve upon strong model-free RL approaches, even when given the additional semantic labels as input to the policy.\todo{Yi: add more texts}
}

Our extensive experiments show that our LEAPS framework outperforms strong model-free RL approaches, even when  the semantic signals are given as input to the policy. Furthermore, the relative improvements of LEAPS over baselines become more significant when the targets are further away from the agent's birthplace, indicating the effectiveness of planning on the learned semantic model.

\section{Related Work}
\label{sec:related}
%\todo{Yi: TODO: move model-based RL in the first place and defer navigation to the end. reminder: to cite A.Gupta's recent navigation works.}

%%%%%%%%%%%%%%%%%%%%%%%%%%%%%%%%
% Generalization
%%%%%%%%%%%%%%%%%%%%%%%%%%%%%%%%
Most deep RL agents are tested in the same training environments~\cite{mirowski2016learning}, disregarding generalization. While limited, robust training approaches have been proposed to enforce an agent's generalization ability, such as domain randomization~\cite{tobin2017domain} and data augmentation by generating random mazes for training~\cite{oh2017zero, parisotto2017neural}. In our work, we use a test set of novel unseen environments, where an agent cannot resort to memorization or simple pattern matching to solve the task.

Meta-learning has shown promising results for fast adaptation to novel environments. Methods include learning a good initialization for gradient descent~\cite{finn2017model} or learning a neural network that can adapt its policy during exploration~\cite{duan2016rl,mishra2017meta}.
We propose to learn a Bayesian model over the semantic level and infer the posterior structure via the Bayes rule. Our approach (1) can work even without any exploration steps in a new environment and (2) is interpretable and can be potentially combined with any graph-based planning algorithm.

Our work can be viewed as a special case of hierarchical reinforcement learning (HRL). Unlike other approaches~\cite{vezhnevets2017feudal,bacon2017option}, in our work high-level planning is performed based on the semantic signals. With orders of magnitudes fewer parameters, our approach is easier to learn compared to recurrent controllers.

LEAPS assumes a discrete semantic signal in addition to the continuous state. A similar assumption is also adopted in~\cite{zhang2018composable}, where the discrete signals are called ``attributes'' and used for planning to solve compositional tasks within the same fully observable environment. \citep{riedmiller2018learning} use additional discrete signals to tackle the sparse reward problem. The schema network~\cite{kansky2017schema} further assumes that even the continuous visual signal can be completely represented in a binary form and therefore directly runs logical reasoning on the binary states.

%%%%%%%%%%%%%%%%%%%%%%%%%%%%%%%%%%%%%%%%%%%%%%%%
% Navigation
%%%%%%%%%%%%%%%%%%%%%%%%%%%%%%%%%%%%%%%%%%%%%%%%

For evaluating our approach, we focus on the problem of visual navigation, which has been studied extensively~\cite{leonard92}. Classical approaches build a 3D map of the scene using SLAM, which is subsequently used for planning~\cite{fox2005}. More recently, end-to-end approaches have been applied to tackle various domains, such  as maze~\cite{mirowski2016learning}, indoor scenes~\cite{zhu2017target} and Google street view~\cite{mirowski2018learning}.
Evidently, navigation performance deteriorates as the agent's distance from the target increases~\cite{zhu2017target,wu2018building}. To aid navigation and boost performance, auxiliary tasks~\cite{mirowski2016learning, jaderberg2016reinforcement} are often introduced during training. Another direction for visual navigation is to use a recurrent neural network and represent the memory in the form of a 2D spatial map~\cite{khan2018memory,parisotto2017neural,tamar2016value,gupta2017cognitive} such that a differentiable planning computation can be performed on the spatial memory. Our approach considers more general graph structures beyond dense 2D grids and captures relationships between \emph{semantic} signals, which we utilize as an informative latent structure in semantically rich environments like House3D.

Similar to our work, Savinov~\etal~\citep{savinov2018graph} constructs a graph of nodes corresponding to different locations of the environment. However, they rely on a pre-exploration step within the test scene and build the graph completely from the pixel space. In LEAPS, we use semantic knowledge and learn a prior over the semantic structures that  are shared across real-world scenes. This allows us to directly start solving for the task at hand without any exploratory steps.

\vspace{-0.5em}
\section{Background}
\vspace{-0.5em}
\label{sec:back}

\hide{
We consider the task of multi-target navigation. The agent is given one of $K$ different targets (concepts), $\mathcal{T}=\{T_1, T_2, \ldots, T_K\}$, and is asked to find an instance of that target concept in a 3D environment (e.g., predefined room types, such as kitchen or bedroom, from the RoomNav task~\cite{wu2018building}).
}

We assume familiarity with standard DRL notations. Complete definitions are in Appendix~\ref{app:define}.

\textbf{Environment:} 
We consider a \emph{contextual Markov decision process}~\cite{hallak2015contextual}  $E(c)$ defined by  $E(c)=(\mathcal{S},\mathcal{A},P(s'|s,a;c),r(s,a;c))$. Here $c$ represents the objects, layouts and any other \emph{semantic} information describing the environment, and is sampled from $\mathcal{C}$, the distribution of possible semantic scenarios.  For  example, $c$ can be intuitively understood as encoding the complete map for navigation, or the complete object and obstacle layouts in robotics manipulations, not known to the agent in advance, and we refer to them as the \emph{context}.

\textbf{Semantic Signal:} At each time step, the agent observes from $s$ a tuple $(s_o, s_s)$, which consists of: (1) a high-dimensional observation $s_o$, e.g., the first person view image, and (2) a low-dimensional  discrete \emph{semantic signal} $s_s$, which encodes semantic information. Such signals are common in AI, e.g., in robotic manipulation tasks $s_s$ indicates whether the robot is holding an object; for games it is the game status of a player; in visual navigation it indicates whether the agent reached a landmark; while in the AI planning literature, $s_s$ is typically a list of \emph{predicates} that describe binary properties of objects. We assume $s_s$ is provided by an oracle function, which can either be directly provided by the environment or extracted by some semantic extractor.

\textbf{Generalization:}
Let $\mu(a|\{s^{(t)}\}_{t};\theta)$ denote the agent's policy parametrized by $\theta$ conditioned on the previous states $\{s^{(t)}\}_t$. The objective of \emph{generalization} is to train a policy on training environments $\mathcal{E}_{\textrm{train}}$ such that the accumulative reward $R(\mu(\theta);c)$ on test set $\mathcal{E}_{\textrm{test}}$ is maximized.

\section{Learning and Planning with a Semantic Model}
\label{sec:graph}

The key motivation of LEAPS is the fact that while each environment can be different in visual appearances, there are structural similarities between environments that can be captured as a probabilistic graphical model over the semantic information. 
On a high level, we aim to learn a Bayesian model $\mathbf{M}^\star(\mathcal{D}, c)$ that captures the semantic properties of the context $c$, from the agent's exploration experiences $\mathcal{D}$. Given a new environment $E(c')$, the agent computes the posterior $P(c'|\mathcal{D}',\mathbf{M}^\star)$ for the unknown context $c'$ via the learned model $\mathbf{M}^\star$ and its current experiences $\mathcal{D}'$. This allows the agent to plan according to its belief of $c'$ to reach the goal more effectively. Thanks to  the Bayes rule, this formulation allows probabilistic inference even with \emph{limited} (or even no) exploration experiences.

Learning an accurate and complete Bayesian model $\mathbf{M}^\star(\mathcal{D}, c)$ can be challenging. We learn an \emph{approximate} latent variable model $\mathbf{M}(y, z; \psi)$ parameterized by $\psi$ with observation variable $y$ and latent variable $z$ that \emph{only} depend on the \emph{semantic signal} $s_s$.
%\hide{  without using the high-dimensional image observation $s_o$. Particularly, we represent our model $\mathbf{M}$ as a \emph{probabilistic graphical model} over the semantic signal $s_s$. }
Suppose we have $K$ different semantic signals $T_1,\ldots,T_K$ and $s_s\in\{0,1\}^{K}$ where $s_s(T_k)$ denotes whether the $k$th signal $T_k$ (e.g., landmarks in navigation) is reached or not. Assuming $T_1$ is the final goal of the task, from any state $s$, we want to reach some final state $s'$ with $s'_s(T_1)=1$. 
In this work, we consider navigation as a concrete example, which can be represented as reaching a state where a desired semantic signal becomes `true'. We exploit the fact that navigation to a target can be decomposed into reaching several way points on way to the target, and therefore can be guided by planning on the semantic signals, i.e., arrival at particular way points. 

\subsection{The Semantic Model}

Note that there can be $2^K$ different values for $s_s$. For efficient computation, we assume \emph{independence} between different semantic signals $T_k$: we use a binary variable $z_{i,j}$ to denote whether some state $s'$ with $s_s'(T_j)=1$ can be ``\emph{directly reached}'', i.e., by a few exploration steps, from some state $s$  with $s_s(T_i)=1$, \emph{regardless} of other signals $T_k\not\in\{T_i,T_j\}$. In addition, we also assume \emph{reversibility}, i.e., $z_{i,j}=z_{j,i}$, so only $K(K-1)/2$ latent variables are needed. 
%%AT: would be good to give some intuition why this is reasonable. Is there a similar relaxation in the planning literature? Or, we can say something like: even though this is an approximation, note that the plan will be refined online through learning, so the approximation made by the model will not limit the final performance of the agent.
%some parameters to be learned from training environments
Before entering the unknown environment, the agent does not know the true value of $z_{i,j}$, but holds some prior belief $P(z_{i,j})$, defined by $z_{i,j}\sim\mathrm{Bernoulli}(\psi^{\mathrm{prior}}_{i,j})$, where $\psi_{i,j}^\mathrm{prior}$ is some parameter to be learned. After some exploration steps, the agent receives a noisy observation $y_{i,j}$ of $z_{i,j}$, i.e., whether a state $s'$ with $s'_s(T_j)=1$ is reached. We define the observation model $P(y_{i,j}|z_{i,j})$ as follows:
\begin{equation}
y_{i,j} \sim \left\{
\begin{array}{ll}
\textrm{Bernoulli}(\psi^\mathrm{obs}_{i,j,0}) & \textrm{ if } z_{i,j}=0\\
\textrm{Bernoulli}(1 - \psi^\mathrm{obs}_{i,j,1}) & \textrm{ if } z_{i,j}=1
\end{array}
\right.
\label{eq:observ}
\end{equation}
At any time step, the agent hold an overall belief $P(z|\mathcal{Y})$ of the semantic structure of the unknown environment, based on its experiences $\mathcal{Y}$, namely the samples of $y$. 

\label{sec:pgm}

\subsection{Combining the Semantic Model with Multi-Target Sub-policies}
%\vspace{-0.1in}
\textbf{Multi-target sub-policies:} With our semantic model, we correspondingly learn multi-target sub-policies $\mu(a|\{s_o^{(t)}\}_t;T_i,\theta)$ taking $s_o$ as input such that $\mu(T_i,\theta)$ is particularly trained for sub-target $T_i$, i.e., reaching a state $s'$ with $s'_s(T_i)=1$. Hence the semantic model can be treated as a model-based \emph{planning module} that picks an intermediate sub-target for the sub-policies to execute so that the final target $T_1$ can be reached with the highest probability. 
Learning the multi-target sub-policies can be accomplished by any standard deep RL method on $\mathcal{E}_{\textrm{train}}$. 

\hide{Now we focus on how to obtain samples to the observation variable $y$ in $\mathbf{M}$ from the semantic signal $s_s$ during the execution of sub-policies.}

%With the help of $s_s$, we update the posterior of our Bayesian model. 
\textbf{Inference and planning on $\mathbf{M}$: }We assume the agent explores the current environment for a \emph{short} horizon of $N$ steps and receives semantic signals ${s_s}^{(1)},\ldots,{s_s}^{(N)}$. Then we compute the bit-OR operation over these binary vectors 
$
B={s_s}^{(1)}\texttt{ OR }\ldots \texttt{ OR }{s_s}^{(N)}.
$
By the reversibility assumption, for $T_i$ and $T_j$ with $B(T_i)=B(T_j)=1$, we know that $T_i$ and $T_j$ are ``directly reachable'' for each other, namely a sample of $y_{i,j}=1$, and otherwise $y_{i,j}=0$. Combining all the history samples of $y$ and the current batch from $B$ as $\mathcal{Y}$, we can perform posterior inference $P(z|\mathcal{Y})$ by the Bayes rule.
By the independence assumption, we can individually compute the belief of each latent variable $z_{i,j}$, denoted by $\hat{z}_{i,j}=P(z_{i,j}|\mathcal{Y}_{i,j})$. Given the current beliefs $\hat{z}_{i,j}$, the current semantic signals $s_s$ and the goal $T_1$, we search for an optimal plan $\tau^* = \{\tau_0, \tau_1, \ldots,\tau_{m-1}, \tau_m\}$, where $\tau_{m}=1$, so that the joint belief along the path from some current signal to the goal is maximized:
\begin{equation}\label{eq:plan}
    \tau^\star = \arg\max_\tau s_s(T_{\tau_0})\prod_{t=1}^m  \hat z_{\tau_{t-1}, \tau_{t}}.
\end{equation}
%%AT: can you add something like: intuitively, in a house navigation task, this is equivalent to searching for the most likely sequence of rooms en route to the goal.
After obtaining $\tau^\star$, we execute the sub-policy for the next sub-target $T_{\tau^\star_1}$, and then repeatedly update the model and replan every $N$ steps.%This is like model predictive control (MPC). 
\label{sec:plan}

\subsection{Learning the Semantic Model}

The model parameters $\psi$ have two parts: $\psi^{\mathrm{prior}}$ for the prior of $z$ and $\psi^{\mathrm{obs}}$ for the noisy observation $y$. Note that $\psi^{\mathrm{obs}}$ is related to the performance of the sub-policies $\mu(\theta)$: if $\mu(\theta)$ has a high success rate for reaching  sub-targets, $\psi^{\mathrm{obs}}$ should be low; when $\mu(\theta)$ is poor, $\psi^{\mathrm{obs}}$ should be higher (cf.~Eq.~\eqref{eq:observ}). %Hence, we propose two objectives to learn these two parts of parameters.

\textbf{Learning $\psi^{\mathrm{prior}}$: } We learn $\psi^{\mathrm{prior}}$ from $\mathcal{E}_{\textrm{train}}$. During training, for each pair of semantic signals $T_i$ and $T_j$, we run \emph{random explorations} from some state $s$ with $s(T_i)=1$. If eventually we reach some state $s'$ with $s'(T_j)=1$, we consider $T_i$ and $T_j$ are reachable and therefore a \emph{positive} sample $z_{i,j}=1$; otherwise a \emph{negative} sample $z_{i,j}=0$. 
Suppose $\mathcal{Z}$ denotes the samples we obtained for $z$ from $\mathcal{E}_{\textrm{train}}$. We run maximum likelihood estimate for $\psi^{\textrm{prior}}$ by maximizing 
$
L_{\textrm{MLE}}(\psi^{\textrm{prior}})=P(\mathcal{Z}|\psi^{\textrm{prior}})
$.

\textbf{Learning $\psi^{\mathrm{obs}}$: } There is no direct supervision for $\psi^{\mathrm{obs}}$. However, we can evaluate a particular value of $\psi^{\mathrm{obs}}$ by policy evaluation on the validation environments $\mathcal{E}_{\textrm{valid}}$. We optimize the accumulative reward 
% $L_{\textrm{valid}}(\psi^{\mathrm{obs}})$ by
$
    L_{\textrm{valid}}(\psi^{\mathrm{obs}})=\mathbb{E}_{E(c)\in\mathcal{E}_{\textrm{valid}}}\left[R(\mu(\theta),M(\psi); c)\right],
$ with the semantic model $M(\psi)$.
%%AT $R$ is previously defined to take in 2 parameters. What does $R$ mean here? (I'm guessing, the policy when using the model $M$ to select targets, but state that explicitly.
Analytically optimizing $L_{\textrm{valid}}$ is hard. Instead, we apply local search in practice to find the optimal $\psi^{\mathrm{obs}}$. 
%%AT: what does local search mean? Is there a specific algorithm you can name or reference? What is the signal for improving $\psi^{\mathrm{obs}}$? This paragraph was very confusing to me. Maybe provide an intuitive explanation for the meaning of $\psi^{\mathrm{obs}}$ and what you're learning here in effect?
%%AT state over what you are searching here.
%\GG{Original text from Yi said we can do either local search or evolutionary method. I believe we did cross validation aka local search.}

\hide{
\textbf{No assumption: } Here we do not assume that we know the true graph representation $z^\star(c)$. Marginalizing over $z$ is typically not tractable. Therefore,  for any particular $\phi$, we can evaluate the effectiveness of $M(;\phi)$ over $\mathcal{E}_{\textrm{train}}$ and find the best $\phi^\star$ leading the best performances. The objective is shown below:
\begin{equation}
    L_{\textrm{meta}}(\phi)=\mathbb{E}_{E(c)\in\mathcal{E}_{\textrm{train}}, T_i}\left[R^\textrm{test}(\mu(M(;\phi));T_i, c)\right]\label{loss:meta}
\end{equation}
where $\mu(M(;\phi))$ is the policy using the guidance from the current model $M(;\phi)$.
A notable point here is, although we are using training environments $\mathcal{E}_{\textrm{train}}$, we should utilize the sparse reward function $R^\textrm{test}$ instead of the shaped dense reward $R^\textrm{train}$ used in training.

Eq.~\ref{loss:meta} can be optimized by local search, greedy algorithm or evolution method~\cite{houthooft2018evolved}.
}

\section{RoomNav: a 3D Navigation Task for RL Generalization}

RoomNav is a concept driven navigation task based on the House3D environment~\cite{wu2018building}. In RoomNav, the agent is given a concept target, i.e., a room type, and needs to navigate to find the target room. RoomNav pre-selected a fixed set of target room types and provides a training set of 200 houses, a testing set  of 50 houses and a small validation set of 20 houses. 

\textbf{Semantic signals: } \hide{The semantic information of a house can be naturally represented as a connectivity graph over rooms. Hence, w} We choose the $K=8$ most common room types as our semantic signals, such that $s_s(T_i)$ denotes whether the agent is currently in a room with type $T_i$\footnote{We also treat $s_s=\mathbf{0}$ as a special  semantic signal. So $\mathbf{M}$ actually contains $K+1$ signals.}. When given a target $T_i$, reaching a state $s$ with $s_s(T_i)=1$ becomes our final goal. House3D provides bounding boxes for rooms, which can be directly used as the oracle for semantic signals. We can also train a room type classifier to extract the semantics signals, which is evaluated in Sec.~\ref{sec:exp_cnn}.

\textbf{The semantic model and sub-policies: } In navigation, the reachability variable $z_{i,j}$ can naturally represent the connectivity between room type $T_i$ and room type $T_j$\footnote{A house can have multiple rooms of the same type. But even this simplification improves generalization.}. We run random explorations in training houses between rooms to collect samples for learning $\psi^{\mathrm{prior}}$. For learning $\psi^{\mathrm{obs}}$, we perform a grid search and evaluate on the validation set. For sub-policies, we learn target driven LSTM policies by A3C~\cite{mnih2016asynchronous} with shaped reward on $\mathcal{E}_{\textrm{train}}$. More details are in Appendix.~\ref{app:policy}.
\hide{
For the neural sub-policy $\mu(\theta)$, we train $K$ different LSTM policies, i.e., $\mu(\theta_i)$ for each target $T_i\in \mathcal{T}$, using . When navigating towards target $T_i$, we simply execute the corresponding policy $\mu(\theta_i)$. We notice that training separate sub-policies leads to better performance than training a multi-target policy as originally described in \cite{wu2018building}.}

\begin{figure}[bt]
\centering
\includegraphics[width=0.31\textwidth]{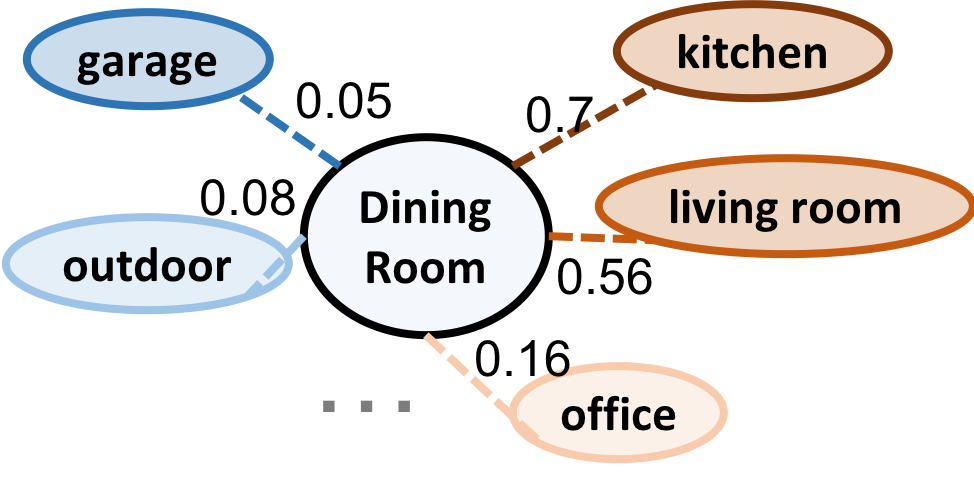}
\includegraphics[width=0.31\textwidth]{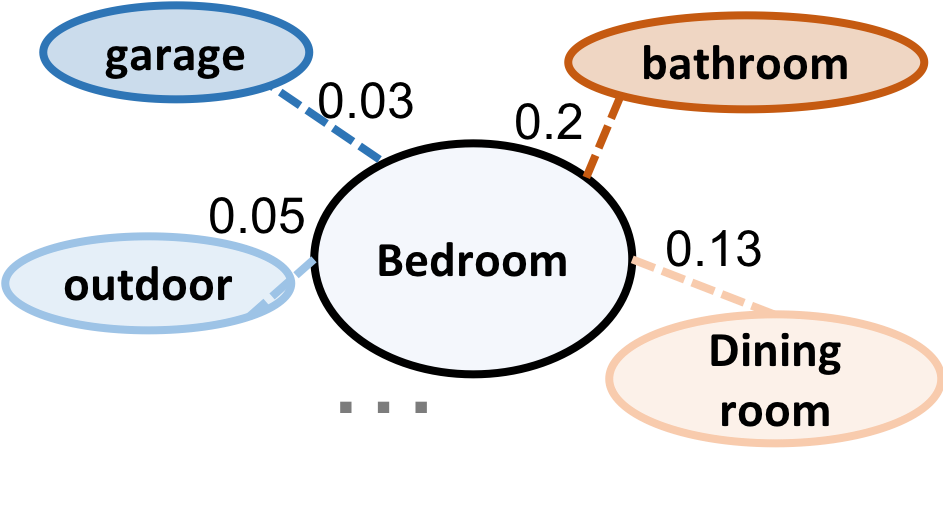}
\includegraphics[width=0.31\textwidth]{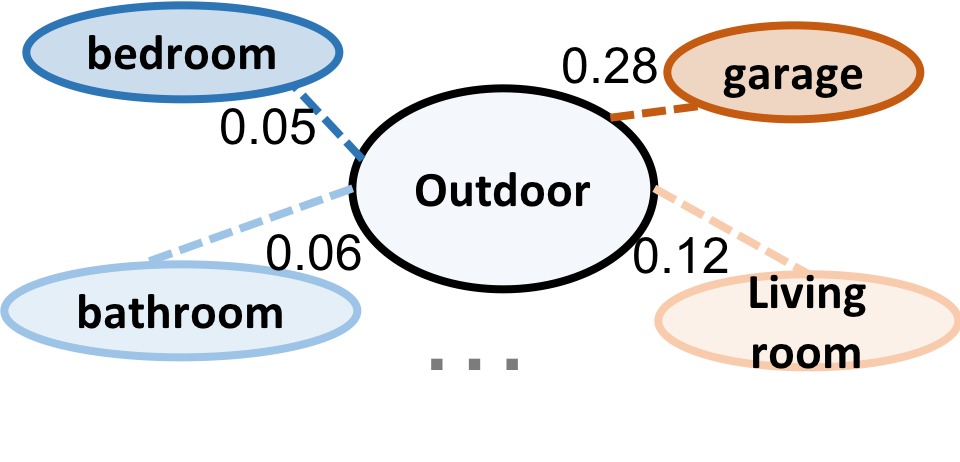}
\vspace{-3mm}
\caption{Visualization of learned semantic prior of $\mathbf{M(\psi)}$: the most and least likely nearby rooms for dining room (L), bedroom (M) and outdoor (R), with numbers denoting $\psi^z$, i.e., the probability of two rooms connecting to each other.\vspace{-2mm}}
\label{fig:graph}
\end{figure}

\begin{figure}[bt!]
    \centering
    \includegraphics[width=0.95\linewidth]{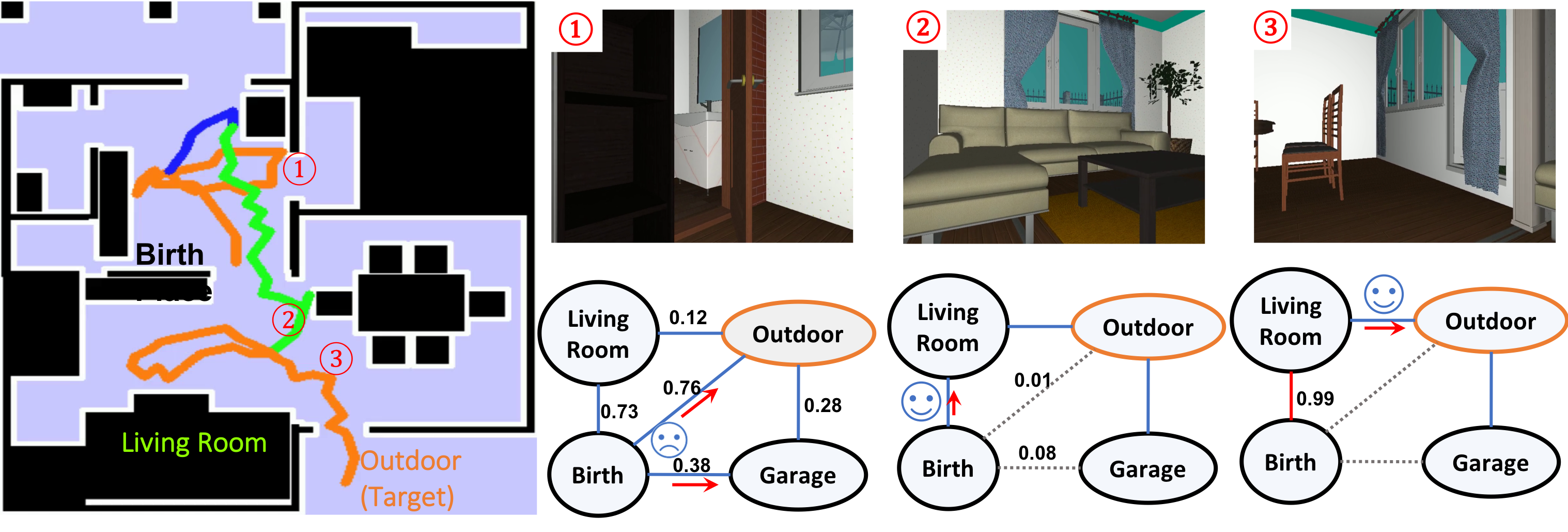}
    \vspace{-3mm}
    \caption{Example of a successful trajectory. The agent is spawned inside the house, targeting ``outdoor''. \textbf{Left}: the 2D top-down map with sub-target trajectories (``outdoor'' -- orange; ``garage'' -- blue; ``living room'' -- green); \textbf{Right, 1st row}: RGB visual image; \textbf{Right, 2nd row}: the posterior of the semantic graph and the proposed sub-targets (red arrow). Initially, the agent starts by executing the sub-policy "outdoor" and then "garage" according to the prior knowledge (\textbf{1st graph}), but both fail (top orange and blue trajectories in the map). After updating its belief that garage and outdoor are not nearby (grey edges in the \textbf{2nd graph}), it then executes the "living room" sub-policy with success (red arrow in the \textbf{2nd graph}, green trajectory). Finally, it executes ``outdoor'' sub-policy again, explores the living room and reaches the goal (\textbf{3rd graph}, bottom orange trajectory).\vspace{-2mm}}
    \label{fig:casestudy_outdoor}
\end{figure}

\section{Experiments}
\label{sec:expr}

In this section, we experiment on RoomNav and try to answer the following questions: \textbf{(1)}  Does the learned prior distribution capture meaningful semantic consistencies? \textbf{(2)} Does our LEAPS agent generalize better than the model-freel RL agent that \emph{only} takes image input? \textbf{(3)}  Our LEAPS agent takes additional semantic signals as input. How does LEAPS compare to other model-free RL approaches that \emph{also} take the semantic signals as part of the inputs but in a different way from our semantic model? For example, what about replacing our semantic model with a complicated RNN controller? \textbf{(4)} House3D provides  labels for rooms (although noisy). \hide{It is not desirable to assume the testing environment can always provide semantic signals.} Can our approach still work if we extract the semantic signals from a trained neural classifier for room types? %We follow \cite{wu2018building} and evaluate the success rate on $\mathcal{E}_{\textrm{test}}$. More details are in appendix.

\subsection{The Learned Prior of the Semantic Model}
We visualize the learned prior $P(z|\psi^{\mathrm{prior}})$ in Fig.~\ref{fig:graph} with 3 room types and their most and least likely connected rooms. The learned prior indeed captures reasonable relationships: bathroom is likely to connect to a bedroom; kitchen is often near a dining room while garage is typically outdoor.

\begin{figure}[bt]
\centering
\includegraphics[width=0.32\textwidth]{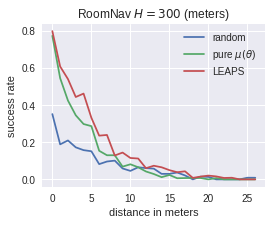}
\includegraphics[width=0.32\textwidth]{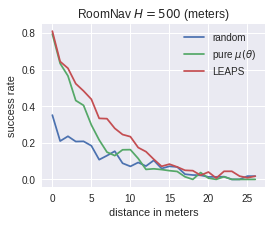}
\includegraphics[width=0.32\textwidth]{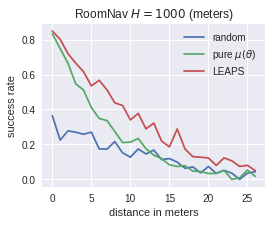}
\includegraphics[width=0.32\textwidth]{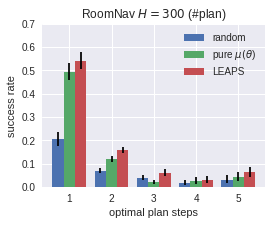}
\includegraphics[width=0.32\textwidth]{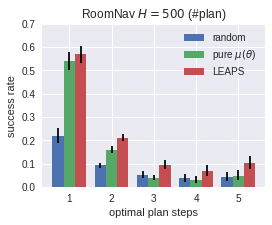}
\includegraphics[width=0.32\textwidth]{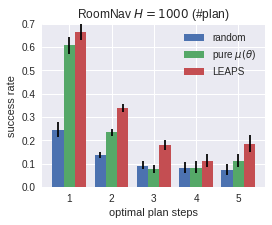}
\includegraphics[width=0.32\textwidth]{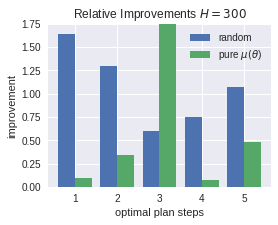}
\includegraphics[width=0.32\textwidth]{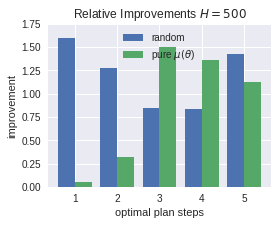}
\includegraphics[width=0.32\textwidth]{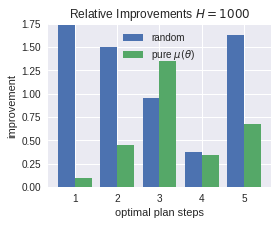}
\vspace{-3mm}
\caption{Comparison with model-free baselines (Sec.~\ref{sec:baseline}). We evaluate performance of random policy (blue), model-free RL baseline (pure $\mu(\theta)$, green) and our LEAPS agent (red), with increasing horizon $H$ from left to right (\textbf{left}: $H=300$; \textbf{middle}: $H=500$; \textbf{right}: $H=1000$). Each row shows a particular metric. \textbf{Top row}: success rate (y-axis) w.r.t. the distance in meters from the birthplace to target room (x-axis); \textbf{middle row}: success rate with confidence interval (y-axis) w.r.t. the shortest planning distance in the ground truth semantic model (x-axis); \textbf{bottom row}: relative improvement of LEAPS over the baseline (y-axis) w.r.t. the optimal plan distance (x-axis). As the number of planning computations, i.e., $H/N$, increases (from left to right), LEAPS agent outperforms baselines more. LEAPS also has higher relative improvements for faraway targets.\vspace{-2mm}}
\label{fig:roomnav}
\end{figure}

\subsection{Comparison with Model-Free RL Baselines}\label{sec:baseline}
We follow the evaluation process in \cite{wu2018building} and measure the testing success rate on $\mathcal{E}_{\textrm{test}}$. More details are in Appendix~\ref{app:eval} and \ref{app:visual}.
We compare our LEAPS agent with two baselines (1) random policy (denoted by ``random'') and (2) model-free RL agent that only takes in image input $s_o$ and executes $\mu(T_i,\theta)$ throughout the episode (denoted by ``pure $\mu(\theta)$''). For LEAPS agent, we set $N=30$, i.e., update the semantic model every 30 steps. We experiment on different horizons $H=300,500,1000$ and evaluate the success rate and relative improvements of our LEAPS agent over the baselines in Fig.~\ref{fig:roomnav}. As the number of planning computations, $H/N$, increases, our LEAPS agent outperforms the baselines more significantly in success rate. Note that the best relative improvements are achieved for targets  neither too faraway nor too close, i.e., optimal plan steps equal to 3 or 4. Interestingly, we observe that there is a small success rate increase for targets that are 5 plan steps away. We suspect that this is because it is rare to see houses that has a diameter of 5 in the semantic model (imagine a house where you need to go through 5 rooms to reach a place). Such houses may have structural properties that makes navigation easier.
Fig.~\ref{fig:casestudy_outdoor} shows an example of a success trajectory of our LEAPS agent. We visualize the progression of the episode, describe the plans and show the updated graph after exploration.

\begin{figure}[bt]
\centering
\includegraphics[width=0.32\textwidth]{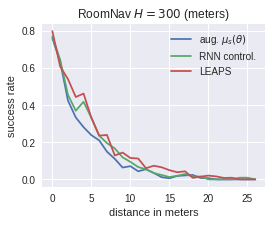}
\includegraphics[width=0.32\textwidth]{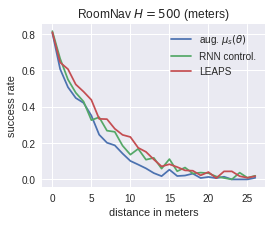}
\includegraphics[width=0.32\textwidth]{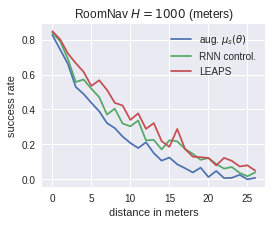}
\includegraphics[width=0.32\textwidth]{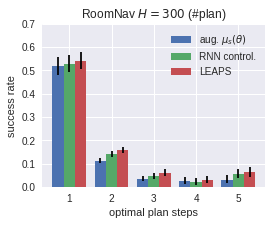}
\includegraphics[width=0.32\textwidth]{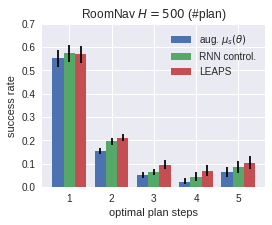}
\includegraphics[width=0.32\textwidth]{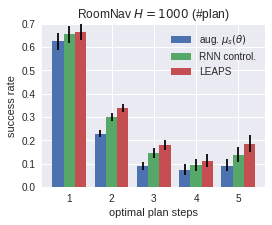}
\includegraphics[width=0.32\textwidth]{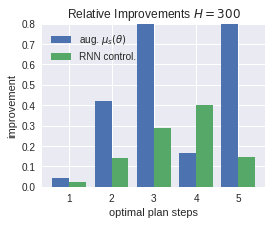}
\includegraphics[width=0.32\textwidth]{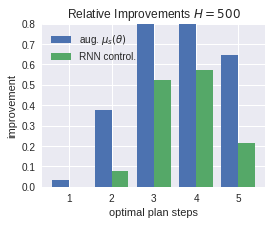}
\includegraphics[width=0.32\textwidth]{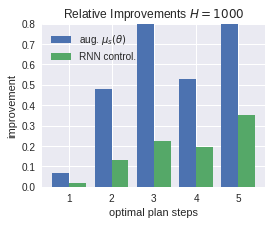}
\vspace{-3mm}
\caption{Comparison with semantic-aware policies (Sec.~\ref{sec:semantic_aware}). We evaluate performance of the semantic augmented model-free agent (``aug. $\mu_s(\theta)$'', blue), the HRL agent with the same sub-policies as LEAPS but with an LSTM controller (``RNN control.'', green) and our LEAPS agent (red), with increasing horizon $H$ from left to right (\textbf{left}: $H=300$; \textbf{middle}: $H=500$; \textbf{right}: $H=1000$). \textbf{Top row}: success rate (y-axis) w.r.t. the distance in meters from birthplace to target (x-axis); \textbf{middle row}: success rate with confidence interval (y-axis) w.r.t. the shortest planning distance in the ground truth semantic model (x-axis); \textbf{bottom row}: relative improvements of LEAPS over the baselines (y-axis) w.r.t. the optimal plan distance (x-axis). Our LEAPS agent outperforms both of the baselines. Note that even though the LSTM controller has two orders of magnitudes more parameters than our semantic model $\mathbf{M}$, our LEAPS agent still performs better, especially for faraway targets.\vspace{-2mm}}\label{fig:baseline}
\end{figure}

\subsection{Comparing to Semantic-Aware Agents without a Graph Representation}\label{sec:semantic_aware}
Here we consider two semantic-aware agents that also takes the semantic signals as input.

\textbf{Semantic augmented agents: } We train new sub-policies $\mu_s(\theta_s)$ taking both $s_o$ and $s_s$ as input.

\textbf{HRL agents with a RNN controller: } Note that updating and planning on $M$ (Eq.~\ref{eq:plan}) only depend on (1) the current semantic signal $s_s$, (2) the target $T_i$, and (3) the accumulative bit-\texttt{OR} feature $B$. Hence, we fixed the same set of sub-policies $\mu(\theta)$ used by our LEAPS agent, and train an LSTM controller with 50 hidden units on $\mathcal{E}_{\textrm{train}}$ that takes all the necessary semantic information, and produce a sub-target every $N$ steps. Training details are in Appendix~\ref{app:baseline}.
Note that the only difference between our LEAPS agent and this HRL agent is the \emph{representation} of the planning module. The LSTM controller has access to exactly the same semantic information as our model $M$ and uses a much more complicated neural model. Thus we expect it to perform competitively to our LEAPS agent. 

The results are shown in Fig.~\ref{fig:baseline}, where our LEAPS agent outperforms both baselines. The semantic augmented policy $\mu_s(\theta_s)$ does not improve much on the original $\mu(\theta)$. For the HRL agent with an LSTM controller, the LEAPS agent achieves higher relative improvements for faraway targets, and also has the following advantages: (1) $M$ can be learned more efficiently with much fewer parameters: an LSTM with 50 hidden units has over $10^4$ parameters while $M(\psi)$ only has 38 parameters\footnote{We assign the same value to all $\psi^{\mathrm{obs}}_{i,j,c}$ for each $c\in\{0,1\}$. See more in Appendix~\ref{app:model}.}; (2) $M$ can adapt to new sub-policies $\mu(\theta')$ with little finetuning ($\psi^\mathrm{prior}$ remains unchanged) while the LSTM controller needs to re-train; (3) the model $M$ and the planning procedure are fully interpretable.

\hide{
\begin{figure}[bt]
\centering
\includegraphics[width=0.33\textwidth]{fig/Baseline_H300.png}
\includegraphics[width=0.32\textwidth]{fig/Baseline_H500.png}
\includegraphics[width=0.32\textwidth]{fig/Baseline_H1000.png}
\caption{Performance within different horizon $H$ on RoomNav of augmented policy $\mu_B(\theta_B)$ (green), RNN controller with $N=30$ (red) and our approach with graphical model $M$ and $N=30$ (purple). X-axis: distance from agent's birth place to target; Y-axis: success rate.\vspace{-2mm}}
\label{fig:baseline}
\end{figure}
}

\begin{figure}[bt]
\centering
\includegraphics[width=0.32\textwidth]{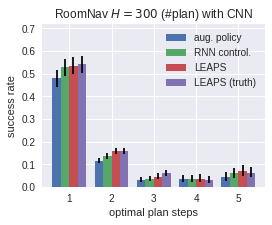}
\includegraphics[width=0.32\textwidth]{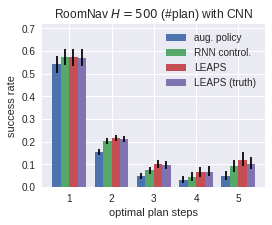}
\includegraphics[width=0.32\textwidth]{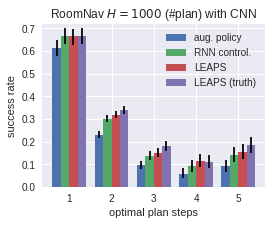}
\includegraphics[width=0.32\textwidth]{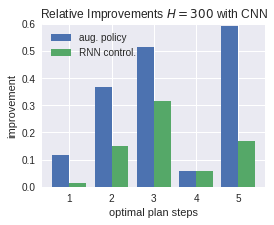}
\includegraphics[width=0.32\textwidth]{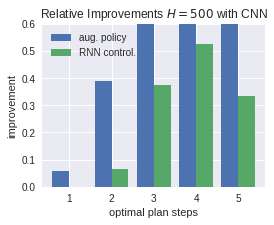}
\includegraphics[width=0.32\textwidth]{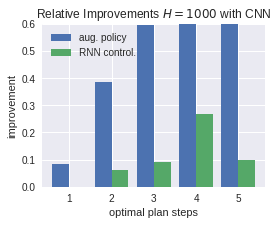}
\vspace{-3mm}
\caption{Performance of semantic-aware agents using a CNN semantic extractor (Sec.~\ref{sec:exp_cnn}). We evaluate performance of the semantic augmented model-free agent (``aug. $\mu_s(\theta)$'', blue), the HRL agent with an LSTM controller (``RNN control.'', green), our LEAPS agent (red) as well as the reference LEAPS agent using ground truth semantic signals (purple), with increasing horizon $H$ from left to right (\textbf{left}: $H=300$; \textbf{middle}: $H=500$; \textbf{right}: $H=1000$). \textbf{Top row}: success rate with confidence interval (y-axis) w.r.t. the shortest planning distance in the ground truth semantic model (x-axis); \textbf{bottom row}: relative improvements of LEAPS over the baselines (y-axis) w.r.t. the optimal plan distance (x-axis). With a CNN extractor, our LEAPS agent still outperforms the baselines, despite a small performance drop compared to the reference LEAPS agent using ground truth signals.\vspace{-2mm}}
\label{fig:classifier}
\end{figure}

\subsection{Using a neural network to extract semantic information}\label{sec:exp_cnn}
In previous experiments, we use the bounding box labels from House3D as the semantic signals. Now we train a CNN room type classifier instead to extract semantic signals from visual input. The CNN classifier are trained on $\mathcal{E}_{\textrm{train}}$ and validated on $\mathcal{E}_{\textrm{valid}}$. More details are in Appendix~\ref{app:cnn}.

We compare our LEAPS agent with the semantic-aware agents using a CNN semantic extractor to produce $s_s$. The results are shown in Fig.~\ref{fig:classifier}, where we also include a reference LEAPS agent using the labels from House3D (purple). Note that all these agents are trained with ground truth semantic signals while only use the CNN extractor during test phase. Our LEAPS agent still outperforms all the baselines and has a comparable performance comparing to the reference LEAPS agent using the ground truth signals.

\section{Conclusion and Future Work}
\label{sec:conclude}
In this work, we proposed LEAPS to improve generalization of RL agents in unseen environments with diverse room layouts and object arrangements, while the underlying semantic information is shared with the environments in which the agent is trained on. We adopt a graphical model over semantic signals, which are low-dimensional binary vectors. During evaluation, starting from a prior obtained from the training set, the agent plans on model, explores the unknown environment, and keeps updating the semantic model after new information arrives. For exploration, sub-policies that focus on multiple targets are pre-trained to execute primitive actions from visual input. The semantic model in LEAPS is lightweight, interpretable and can be updated dynamically with little explorations. 
As illustrated in the House3D environment, LEAPS works well for environments with semantic consistencies -- typical of realistic domains. On random environments, e.g., random mazes, LEAPS degenerates to exhaustive search. 

% We illustrate the proposed model in the House3D environment which has intrinsic semantic structures.  
Our approach is general and can be applied to other tasks, such as robotics manipulations where semantic signals can be status of robot arms and object locations, or video games where we can plan on semantic signals such as the game status or current resources. In future work we will investigate models for more complex semantic structures.
% Our current design of a semantic model is also much simplified for computational purposes such that it only considers independent pairwise semantic relationships. But it can be extended to handle more complicated semantic structures.

\bibliography{iclr2019_conference}
\bibliographystyle{iclr2019_conference}

\newpage

\appendix

\section{Complete Notations and Definitions}\label{app:define}

\textbf{Environment:} 
We consider a \emph{contextual Markov Decision Process}~\cite{hallak2015contextual}  $E(c)$ defined by  $E(c)=(\mathcal{S},\mathcal{A},P(s'|s,a;c),r(s,a;c))$, where $\mathcal{S}$ is the state space and $\mathcal{A}$ is the action space. $c$ represents the objects, layouts and any other \emph{semantic} information describing the environment, and is sampled from $\mathcal{C}$, the distribution of possible semantic scenarios. $r(s,a;c)$ denotes the reward function while $P(s'|s,a;c)$ describes transition probability conditioned on $c$. For  example, $c$ can be intuitively understood as encoding the complete map for navigation, or the complete object and obstacle layouts in robotics manipulations, not known to the agent in advance, and we refer to them as the \emph{context}.

\textbf{Semantic Signal:} At each time step, the agent's observation is a tuple $(s_o, s_s)$, which consists of: (a) a high-dimensional observation $s_o$, \eg the first person view image, and (b) a low-dimensional \emph{semantic signal} $s_s$, which encodes semantic information. Such low-dimensional discrete signals are commonly used in AI, \eg in robotic manipulation tasks $s_s$ indicates whether the robot is holding an object; for games it is the game status of a player; in visual navigation it indicates whether the agent reached a landmark; while in the AI planning literature, $s_s$ is typically a list of \emph{predicates} that describe binary properties of objects. We assume $s_s$ is provided by an oracle function, which can either be directly provided by the environment or extracted by some semantic extractor.

\textbf{Generalization:}
Let $\mu(a|\{s^{(t)}\}_{t};\theta)$ denote the agent's policy parametrized by $\theta$ conditioned on the previous states $\{s^{(t)}\}_t$ and $R(\mu(\theta);c)$ \hide{(short for $\mu(\theta)$)} denote the accumulative reward of $\mu(\theta)$ in $E(c)$. 
The objective is to find the best policy that maximizes the \emph{expected} accumulative reward $\mathbb{E}_{c\sim\mathcal{C}}\left[R(\mu(\theta);c,T_i)\right]$.
In practice, we sample a disjoint partition of a training set $\mathcal{E}_{\textrm{train}}=\{E(c_i)\}_i$ and a testing set $\mathcal{E}_{\textrm{test}}=\{E(c_j)\}_j$, where $\{c_i\}$ and $\{c_j\}$ are samples from $\mathcal{C}$. We train $\mu(\theta)$ with a \emph{shaped} reward $r_{\textrm{train}}$ only on $\mathcal{E}_{\textrm{train}}$,  and measure the \emph{empirical} generalization performance of the learned policy on $\mathcal{E}_{\textrm{test}}$ with the original unshaped reward (e.g., binary reward of success or not).

\section{Environment Details}
In RoomNav the 8 targets are: kitchen, living room, dining room, bedroom, bathroom, office, garage and outdoor.
%For ObjectNav, the 15 object targets are: kitchen cabinet, sofa, chair, toilet, table, sink, wardrobe cabinet, bed, shelving, desk, television, household appliance, dresser, vehicle and pool. 
We inherit the success measure of ``see'' from \cite{wu2018building}: the agent needs to see some corresponding object for at least 450 pixels in the input frame and stay in the target area for at least 3 time steps.

For the binary signal $s_s$, we obtain from the bounding box information for each room provided from SUNCG dataset~\cite{song2016ssc}, which is very noisy.

Originally the House3D environment supports 13 discrete actions. Here we reduce it to 9 actions: large forward, forward, left-forward, right-forward, large left rotate, large right rotate, left rotate, right rotate and stay still.

\section{Evaluation Details}\label{app:eval}

We following the evaluation setup from \cite{wu2018building} and measure the success rate on $\mathcal{E}_{\textrm{test}}$ over 5750 test episodes, which consists of 5000 random generated configurations and 750 specialized for faraway targets to increase the confidence of measured success rate. These 750 episodes are generated such that for each plan-distance, there are at least 500 evaluation episodes. Each test episode has a fixed configuration for a fair comparison between different approaches, i.e., the agent will always start from the same location with the same target in that episode. Note that we always ensure that (1) the target is connected to the birthplace of the agent, and (2) the the birthplace of the agent is never within the target room.

\section{Visualization Details}\label{app:visual}
For confidence interval of the measured success rate, we computed it by fitting a binomial distribution.

For optimal plan steps, we firstly extract all the room locations, and then construct a graph where a vertex is a room while an edge between two vertices is the shortest distance between these two rooms. After obtaining the graph and a birthplace of the agent, we compute shortest path from the birthplace to the target on this graph to derive the optimal plan steps.

\section{Details for Learning Neural Sub-Policies}\label{app:policy}
We utilize the same policy architecture as \cite{wu2018building}. It was mentioned in \cite{wu2018building} that using segmentation mask + depth signals as input leads to relatively better performances for policy learning. So we inherit this setting here.
We run A3C with $\gamma=0.97$, batch size 64, learning rate 0.001 with Adam, weight decay $10^{-5}$, entropy bonus 0.1. We backprop through at most 30 time steps. We also compute the squared $l_2$ norm of logits and added to the loss with a coefficient 0.01. We also normalize the advantage to mean 0 and standard deviation 1.

We run a curriculum learning by increasing the maximum of distance between agent's birth meters and target by 3 meters every 10000 iterations. We totally run 60000 training iterations and use the final model as our learned policy $\mu(\theta)$.

In the original House3D paper, a gated attention module is used to incorporate the target instruction. Here, since we only have $K=8$ different sub-policies, we simply train an individual policy for each target and we empirically observe that this leads to better performances.

\section{Details for Learning the Semantic Model}\label{app:model}
After evalution on the validation set, we choose to run random exploration for 300 steps to collect a sample of $z$. For a particular environment, we collect totally 50 samples for each $z_{i,j}$. 

For all $i\ne j$, we set $\psi^{\mathrm{obs}}_{i,j,0}=0.001$ and $\psi^{\mathrm{obs}}_{i,j,1}=0.15$.

\section{Additional Details for Training Semantic-Aware Policies}\label{app:baseline}
For the LSTM controller, we ran A2C with batch size 32, learning rate 0.001 with adam, weight decay 0.00001, gamma $0.99$, entropy bonus 0.01 and advantage normalization. The reward function is designed as follows: for every subtask it propose, it gets a time penalty of 0.1; when the agent reach the target, it gets a success bonus of 2.

The input of the LSTM controller consists of (1) ${s_s}^{(t)}$ ($K$ bits), (2) $B$ ($K$ bits), (3) last subtask $T_k$, and (4) the final target $T_i$. We convert $T_i$ and $T_k$ to a one-hot vector and combine the other two features to feed into the LSTM. Hence the input dimension of LSTM controller is $4K$, namely 32 in RoomNav.

For the semantic augmented LSTM policy, $\mu_s(\theta_s)$, we firstly use the CNN extract visual features from $s_o$ and combine the input semantic features and the visual features as the combined input to the LSTM in the policy.

\section{Additional Details for Training the CNN Semantic Extractor}\label{app:cnn}
We noticed that in order to have a room type classifier, only using the single first person view image is not enough. For example, the agent may face towards a wall, which is not informative, but is indeed inside the bedroom (and the bed is just behind).

So we take the panoramic view as input, which consists of 4 images, $s_o^{1},\ldots,s_o^{4}$ with different first person view angles. The only exception is that for target ``outdoor'', we notice that instead of using a panoramic view, simply keeping the recent 4 frames in the trajectory leads to the best prediction accuracy. We use an CNN feature extractor to extract features $f(s_o^i)$ by applying CNN layers with kernel size 3, strides $[1, 1, 1,  2,  1, 2, 1, 2, 1, 2]$ and channels $[4, 8, 16, 16, 32, 32, 64, 64, 128, 256]$. We also use relu activation and batch norm. 
Then we compute the attention weights over these 4 visual features by $l_i=f(s_o^i)W_1^TW_2\left[f(s_o^1),\ldots,f(s_o^4)\right]$ and $a_i=\textrm{softmax}(l_i)$. Then we compute the weighted average of these four frames $g=\sum_i a_i f(s_o^i)$ and feed it to a single layer perceptron with 32 hidden units. For each semantic signal, we generate 15k positive and 15k negative training data from $\mathcal{E}_{\textrm{train}}$ and use Adam optimizer with learning rate $5e$-$4$, weight decay $1e$-$5$, batch size 256 and gradient clip of 5. We keep the model that has the best prediction accuracy on $\mathcal{E}_{\textrm{valid}}$.

For a smooth prediction during testing, we also have a hard threshold and filtering process on the CNN outputs: $s_s(T_i)$ will be 1 only if the output of CNN has confidence over 0.85 for consecutively 3 steps.

%\textbf{Complete learned graph for rooms and objects: }

%\textbf{Details for learning RNN controller: }

%\textbf{Averaged Success Rates: }

\end{document}